\pdfoutput=1

\documentclass[11pt]{article}

\usepackage[preprint]{acl}
\usepackage{amssymb}
\usepackage{times}
\usepackage{latexsym}

\usepackage[T1]{fontenc}

\usepackage[utf8]{inputenc}

\usepackage{microtype}

\usepackage{inconsolata}

\usepackage{graphicx}
\usepackage{subcaption}
\usepackage{enumitem}
\usepackage{amsmath}  
\usepackage{natbib}
\title{How Syntax Specialization Emerges in Language Models}

\author{
  \textbf{Xufeng Duan\textsuperscript{1}},
  \textbf{Zhaoqian Yao\textsuperscript{1}},
  \textbf{Yunhao Zhang\textsuperscript{2}},\\
  \textbf{Shaonan Wang\textsuperscript{2}},
  \textbf{Zhenguang G. Cai\textsuperscript{1,3}}
\\
  \textsuperscript{1}Department of Linguistics and Modern Languages, The Chinese University of Hong Kong\\
  \textsuperscript{2}Institute of Automation, Chinese Academy of Sciences\\
  \textsuperscript{3}Brain and Mind Institute, The Chinese University of Hong Kong
\\
  \small{
    \href{mailto:xufeng.duan@link.cuhk.edu.hk}{xufeng.duan@link.cuhk.edu.hk}
 }
 \small{
    \href{mailto:zhangyunhao2021@ia.ac.cn}
    {zhangyunhao2021@ia.ac.cn}
 }
}
\usepackage{hyperref}

\begin{document}
\maketitle
\begin{abstract}
Large language models (LLMs) have been found to develop surprising internal specializations: Individual neurons, attention heads, and circuits become selectively sensitive to syntactic structure, reflecting patterns observed in the human brain. While this specialization is well-documented, how it emerges during training—and what influences its development—remains largely unknown.
In this work, we tap into the black box of specialization by tracking its formation over time. By quantifying internal syntactic consistency across minimal pairs from various syntactic phenomena, we identify a clear developmental trajectory: Syntactic sensitivity emerges gradually, concentrates in specific layers, and exhibits a "critical period" of rapid internal specialization. This process is consistent across architectures and initialization parameters (e.g., random seeds), and is influenced by model scale and training data. We therefore reveal not only where syntax arises in LLMs but also how some models internalize it during training. To support future research, we will release the code, models, and training checkpoints upon acceptance.

\end{abstract}

\section{Introduction}
Human language processing is deeply rooted in neural specialization. Decades of neurocognitive research have revealed that regions such as Broca’s and Wernicke’s areas play critical roles in syntactic parsing and linguistic comprehension \cite{friederici2018neural, klein2023children}. Furthermore, neurodevelopmental evidence points to a critical period during which language related knowledge is most effectively acquired, underscoring the structured, staged nature of language learning in the brain \cite{Keshavan2014}.

Intriguingly, recent studies suggest that large language models (LLMs) may exhibit analogous properties. Despite being trained solely on text, LLMs appear to develop internal representations that not only support syntactic processing—such as grammaticality judgments \cite{linzen-etal-2016-assessing}, and structure-sensitive generation \cite{marvin2018targeted}—but also display emergent structural organization reminiscent of brain linguistic specialization: Prior work has identified components such as syntax-sensitive neurons \cite{brinkmann2025large,tang2024language}, attention heads \cite{clark-etal-2019-bert, olsson2022context}, and circuits \cite{ameisen2025circuit,wang2022interpretability} using interpretability techniques that correlate internal model components with syntactic features. However, these approaches offer only limited insight into the developmental dynamics of this specialization within the model. Consequently, we still lack a mechanistic account of how language knowledge (e.g., syntax) \textbf{emerges} during training, and how training related factors influence this process. To address this gap, we ask:

\textbf{RQ1.} How does syntactic specialization emerge and evolve over training?

\textbf{RQ2.} How do model initialization, scale, and training data, shape this development?

\begin{figure*}[htbp]
\centering
\begin{subfigure}[b]{0.48\textwidth}
\includegraphics[width=\textwidth]{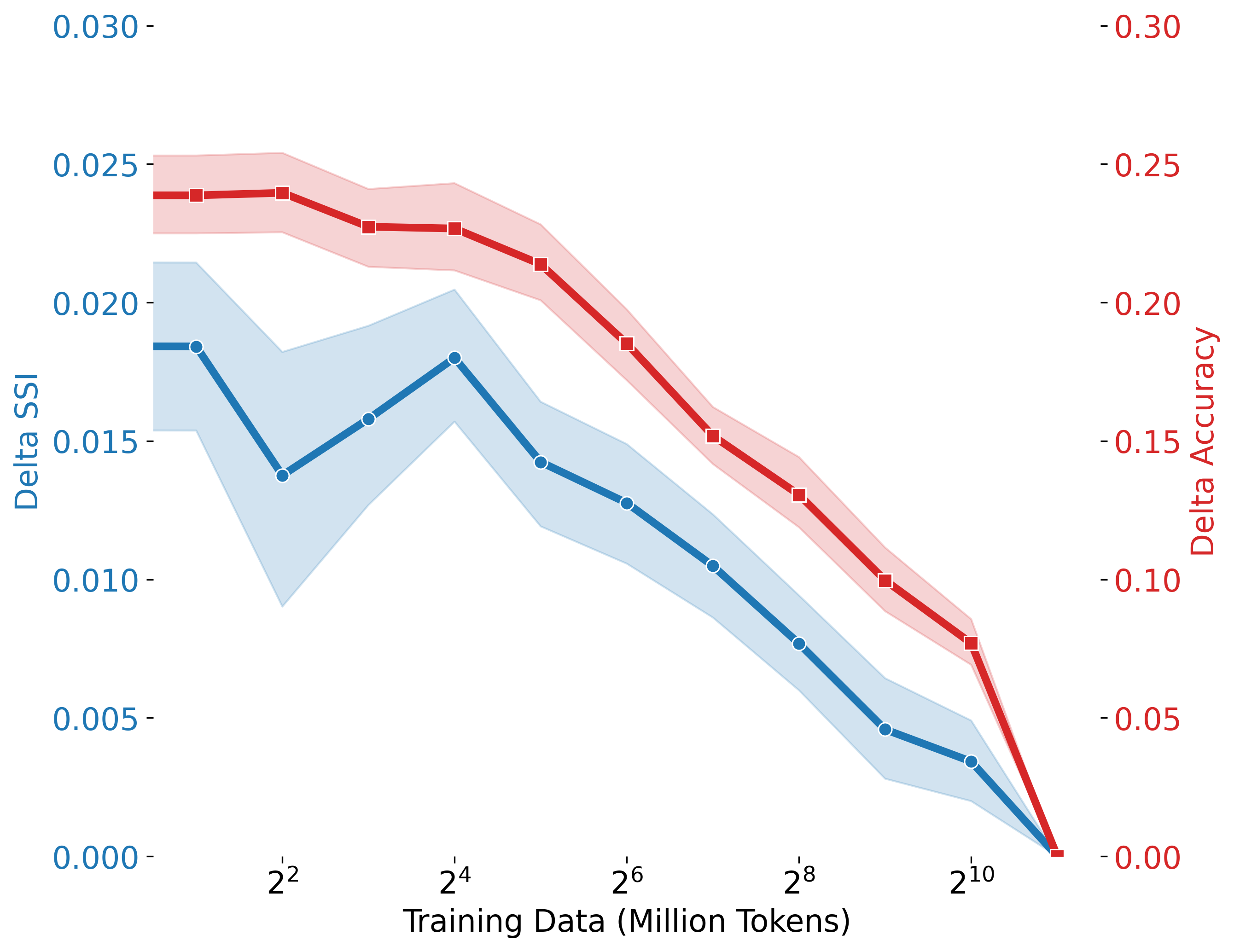}
\end{subfigure}
\hfill
\begin{subfigure}[b]{0.48\textwidth}
\includegraphics[width=\textwidth]{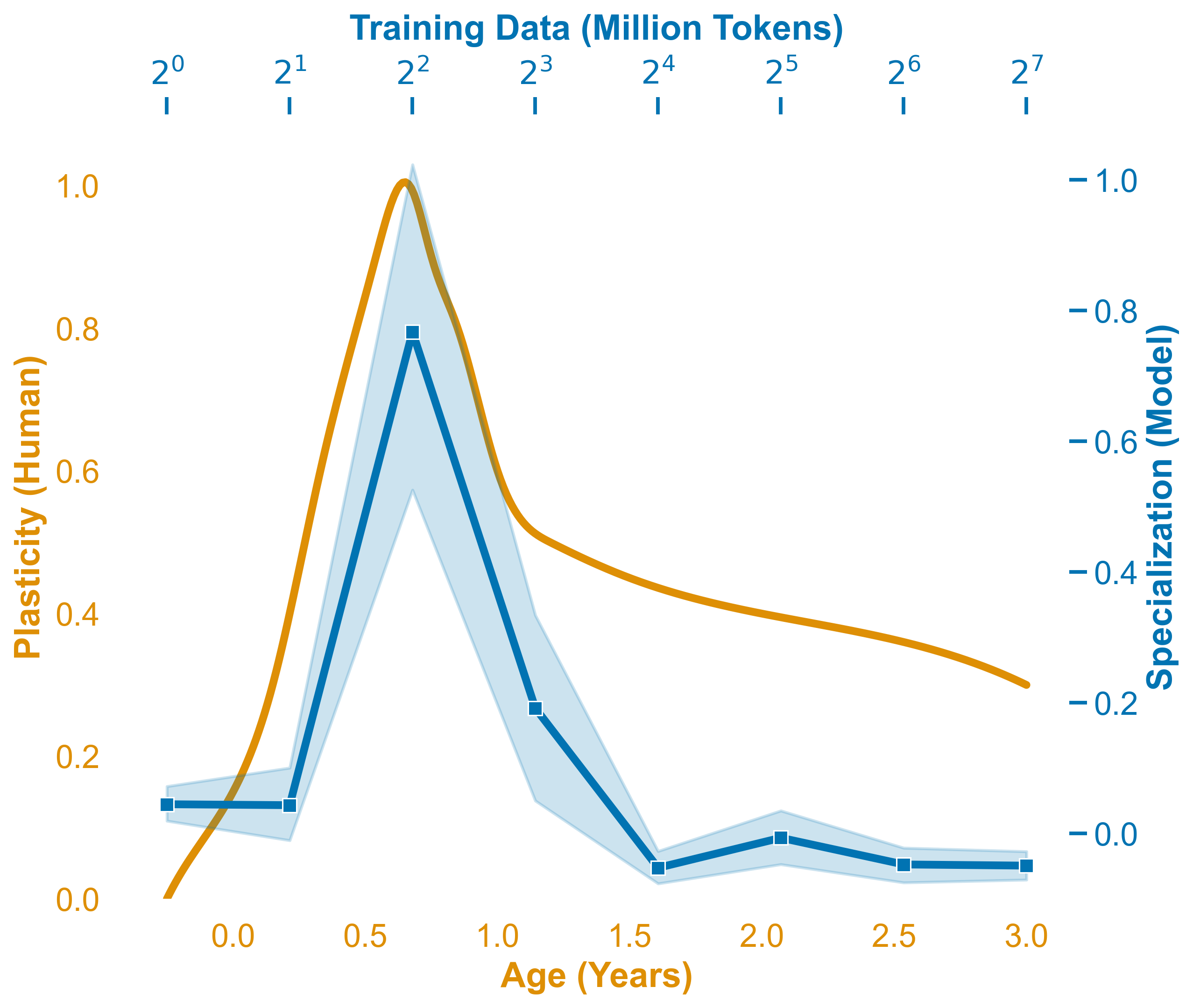}
\end{subfigure}
\caption{(a) Left. Differences in SSI and grammatical judgment task accuracy between various checkpoints during training and the final model. The correlated downward trends suggest that increased syntactic specialization (lower $\Delta$SSI) is associated with improved grammatical accuracy (lower $\Delta$Accuracy). (b) Right. Human neural plasticity (orange) peaks early in development, consistent with the critical period for language acquisition \cite{Keshavan2014}. Analogously, model syntactic specialization (blue), measured by SSI divergence across seeds (see \ref{criticalperiod}), emerges early in training and converges after approximately 16 million tokens.}
\label{fig:image1a}
\label{fig:image1b}
\label{fig:image1}
\end{figure*}

To address these questions, we introduce a novel metric—the Syntactic Sensitivity Index (SSI)—designed to measure syntax specialization by assessing how consistently individual neurons or layers distinguish between grammatical and ungrammatical sentences across diverse syntactic phenomena. High SSI scores indicate a higher degree of specialization. Unlike conventional probing methods (e.g., SVM, regression), SSI does not rely on training classifiers or task-specific supervision; instead, it intrinsically captures the structural separation of syntax-sensitive activations within a model, avoiding shortcomings of supervised learning such as reliance on shallow heuristics or irrelevant features. Using this method, we reveal that:

\textbf{The efficacy of SSI:} During training, layer-wise SSI increases alongside syntax acquisition (see Figure~\ref{fig:image1a} a); SSI significantly predicts the accuracy in grammaticality judgment task, outperforming SVM and regression baselines); ablating high-SSI neurons degrades the sentence’s perplexity (PPL).

\textbf{The development of syntactic specialization:} SSI increased gradually during training and is modulated by the size of training data and model scale.

\textbf{The critical period for syntactic specialization:} Models initialized with different random seeds converge on similar layer-level syntactic representations after approximately 16 million tokens of training, suggesting a critical period during which specialization rapidly emerges (see Figure~\ref{fig:image1} b).

\textbf{Convergent layers, divergent neurons:} While syntactic specialization consistently emerges at the layer level, the specific neurons involved differ across models with different random seeds.

Together, these findings offer a mechanistic account of \emph{when}, \emph{where}, and \emph{how} LLMs specialize syntactic structure.

\section{Related Work}
\subsection{Neural Specialization and Plasticity in Human Language Acquisition}

\begin{figure}[h]
\centering
\includegraphics[width=\columnwidth]{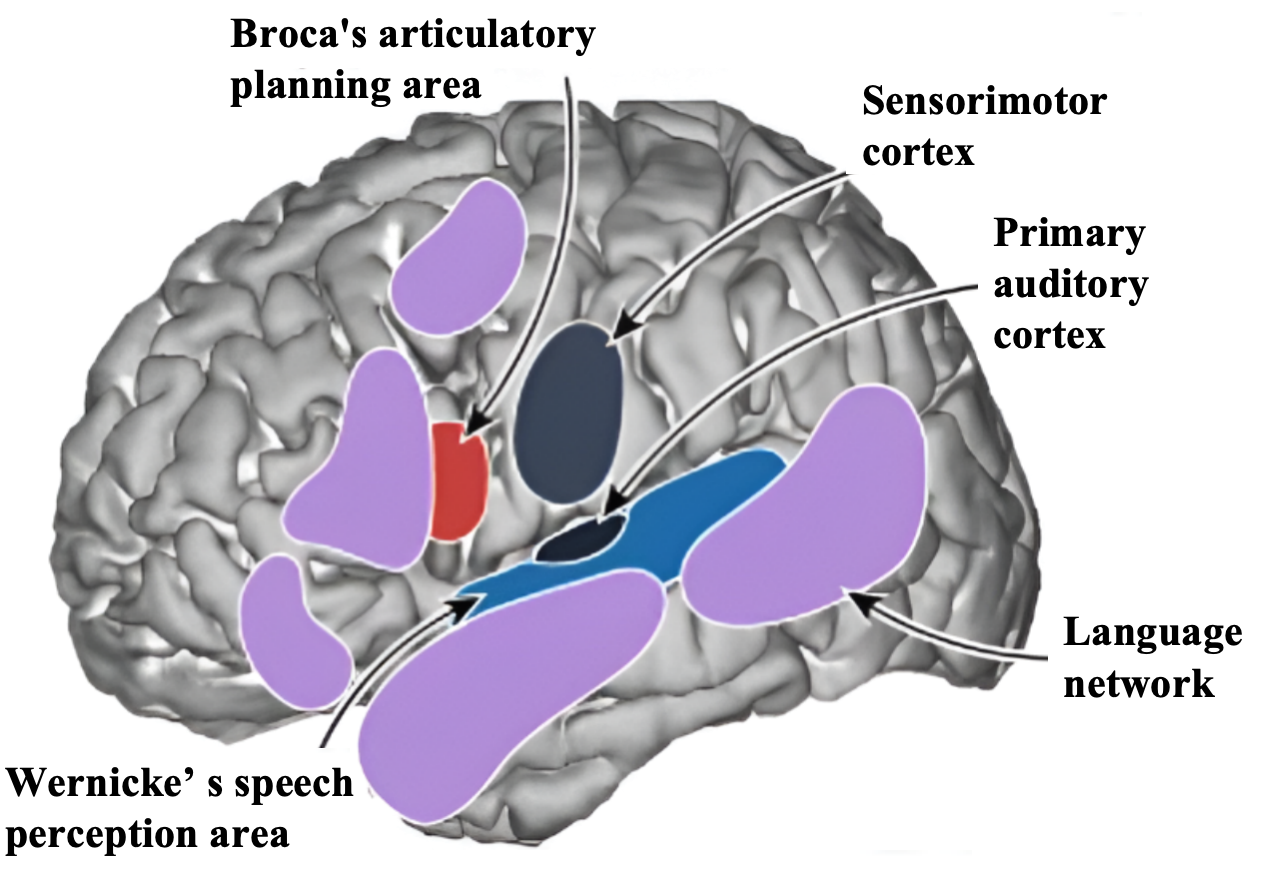}
\caption{Specialized language related regions from \cite{fedorenko2024language}. Broca’s area (red) is implicated in syntactic parsing and articulatory planning, while Wernicke’s area (light blue) supports speech sound processing. Additional regions, including the sensorimotor cortex (dark blue), primary auditory cortex (black), and the broader language network (purple), contribute to the perception, production, and comprehension of language.}
\label{fig:image12}
\end{figure}

Human language acquisition is supported by dedicated brain areas (see Figure~\ref{fig:image12}), most notably in the left perisylvian cortex—Broca's area (syntax and production) and Wernicke's area (comprehension). Lesion and fMRI studies show these regions respond selectively to linguistic input, independent of other cognitive functions \cite{fedorenko2009,fedorenko2024language}. This specialization emerges early in life and aligns with the critical period hypothesis, which posits a biologically constrained window for acquiring syntax and grammar \cite{lenneberg1967,johnson1989}. The developmental trajectory of language-related neuroplasticity (see Figure~\ref{fig:image1}) demonstrates a marked rise beginning shortly after birth, peaking around the end of the first year, and gradually declining thereafter. This pattern reflects a critical window during early infancy when experience-expectant neuroplasticity facilitates rapid synaptogenesis in brain regions responsible for language acquisition \cite{Keshavan2014}.

Empirical evidence from late first-language learners (e.g., deaf individuals exposed to sign language only after childhood) shows persistent syntactic deficits despite extensive use, supporting the idea that grammar learning depends on early neural plasticity \cite{hall2018,penicaud2013}. Similarly, cases like ``Genie'' and ASL learners exposed post-puberty reveal profound limitations in syntactic development \cite{curtiss1977}. Children with early left-hemisphere injuries can often reorganize language to the right hemisphere, unlike adults with similar damage \cite{vicari2000,hertz2002}. These findings highlight a sensitive period in which both language acquisition and cortical specialization are most flexible.

\subsection{Probing Syntactic Competence and Specialization}
Probing studies demonstrate that LLMs capture hierarchical grammatical patterns. \citet{linzen-etal-2016-assessing} showed that LSTMs enforce subject–verb agreement over intervening material. \citet{marvin2018targeted}, and later \citet{warstadt-etal-2020-blimp-benchmark}, extended this to broader syntax via BLiMP—a suite of 67 minimal-pair tests targeting phenomena like reflexive binding, wh-islands, and polarity licensing. Transformer-based models like BERT and GPT perform near human levels on BLiMP, indicating substantial syntactic generalization \cite{linzen-etal-2016-assessing,marvin2018targeted,warstadt-etal-2020-blimp-benchmark}.

Interpretability work has revealed syntax-sensitive components inside LLMs \cite{mueller2022causal, duan2025unveiling, liu2025brain,tang2024language}. \citet{lakretz2019} identified number-agreement neurons in LSTMs. \citet{mueller2022causal} used causal mediation to find MLP neurons in GPT-2 that control agreement decisions. Attention-based studies \cite{clark-etal-2019-bert,voita-etal-2019-analyzing} found syntactic attention heads that consistently track grammatical relations like subject–verb dependencies or reflexive antecedents. \citet{chen2024sudden} formalized this as ``Syntactic Attention Structure,'' showing heads that align with parse structures emerge without supervision. \citet{alkhamissi2025language} found that syntactic (formal) representations in LLMs align more closely with human brain activity than semantic (functional) ones (e.g., world knowledge), suggesting a convergent representational structure.

\section{Method}
\subsection{Syntactic Sensitivity Index (SSI)}
To quantify syntactic specialization in language models, we introduce the SSI—a metric that captures how consistently the model distinguishes grammatical from ungrammatical inputs across specific syntactic phenomena (for input sentences, see \ref{minimal pairs}).

For each minimal pair (a grammatical sentence and its ungrammatical counterpart), we compute the activation difference at each layer:
\[
h_g = Model(grammatical sentence)
\]
\[  
h_u = Model(ungrammatical sentence)
\]
\[
    \Delta h = h_g - h_u
\]

We use mean-pooled sentence embeddings for each layer normalized by their $L_2$ norm, to compute $\Delta h$.

Next, for each syntactic phenomenon $p$ at each layer $l$, we compute: Intra-group and Inter-group Similarity, inspired by previous works \citep{raghu2017svcca,zeng-etal-2025-converging}, which assess similarity between representations.

\textbf{Intra-group similarity}: The average cosine similarity between all activation differences $\Delta h$ from sentence pairs belonging to the same phenomenon.

\[
\textrm{Intra}_p^{(l)} = \frac{1}{|S_p|(|S_p| - 1)} \sum_{\substack{i,j \in S_p \\ i \neq j}} \cos\left( \Delta h_i^{(l)}, \Delta h_j^{(l)} \right)
\]

\textbf{Inter-group similarity}: The average cosine similarity between activation differences for phenomenon $p$ and those from all other phenomena $-S_p$.

\[
\text{Inter}_p^{(l)} = \frac{1}{|S_p| \cdot |{-S_p}|} \sum_{\substack{i \in S_p \\ j \in -S_p}} \cos\left( \Delta h_i^{(l)}, \Delta h_j^{(l)} \right)
\]

The \textbf{SSI }is then defined as:
\[
\text{SSI}_p^{(l)} = \text{Intra}_p^{(l)} - \text{Inter}_p^{(l)}
\]

A higher SSI indicates that the model's activation patterns for grammatical versus ungrammatical contrasts within a syntactic phenomenon are more consistent (high intra-group similarity) and more distinct from those of other phenomena (low inter-group similarity). This suggests the model has developed stable, phenomenon-specific internal representations—a key feature of syntactic specialization.

\subsection{Neuron-Level Specialization via SSI}
\label{neuron-level SSI}
We define a `neuron' as a scalar activation in the feedforward sub-layer (MLP) of a transformer block---that is, one of the dimensions in the model's hidden representation. To localize syntactic sensitivity beyond layer-level aggregates, we extend SSI analysis to the neuron (activation dimension) level. For each neuron $d$, we extract its scalar value in the activation difference $\Delta h$ across all samples of a syntactic phenomenon. We then compute:

\textbf{Within-phenomenon correlation}: The average Pearson correlation of neuron $d$'s response across all minimal pairs within a phenomenon.

\textbf{Distinctiveness score}: The $z$-score of neuron $d$'s correlation relative to a background distribution formed from other phenomena.

Neurons falling within the top 25\% correlation and exceeding a $z$-score threshold of 2 are identified as neurons sensitive to the syntactic phenomenon in question \citep{zeng-etal-2025-converging}. This approach allows us to identify specific neurons that consistently contribute to syntactic discrimination.

\subsection{Minimal Pairs and Dataset}
\label{minimal pairs}
We use the BLiMP benchmark \citep{warstadt-etal-2020-blimp-benchmark}, a comprehensive dataset covering 13 syntactic phenomena, each represented by more than 1,000 minimal pairs (67,000 in total)---one grammatical sentence (e.g., ``The author laughs loudly'') and one ungrammatical counterpart (e.g., ``The author laugh loudly'') differing only in the targeted syntactic feature. These pairs provide controlled contrasts across a wide variety of syntactic categories, including subject-verb agreement, reflexive binding, \textit{wh}-islands, and ellipsis, among others. We process each sentence through a model and extract sentence-level representations layer by layer.

\subsection{Training Dynamics and Correlation Analysis}
To investigate the developmental trajectory of syntactic specialization during training, we compute the SSI progression and assess its alignment with the model’s syntactic competence (Accuracy). For each checkpoint, we measure:

$\textbf{SSI progression} = |\text{SSI}_{\text{final}} - \text{SSI}_{\text{checkpoint}}|$

Accuracy is computed using minimal sentence pairs, where each pair consists of a grammatical sentence $g$ and its ungrammatical counterpart $u$ \cite{liu2024zhoblimp}. For each sentence, we calculate its mean log-probability ($\text{MP}$):
\[
\text{MP}(S) = \frac{1}{|S|} \log P(S)
\]
The model is considered correct if it assigns a higher $\text{MP}$ to the grammatical sentence. Accuracy is then defined as:

\[
\text{Accuracy} = \frac{1}{|P|} \sum_{(g, u) \in P} \mathbb{I} \left( \text{MP}(g) > \text{MP}(u) \right)
\]
where $\mathbb{I}$ is the indicator function, returning 1 if the condition is true and 0 otherwise. Also 

\textbf{Accuracy progression}: \\
\indent $\Delta \text{Acc} = |\text{Acc}_{\text{final}} - \text{Acc}_{\text{checkpoint}}|$

We then compute the correlation between $\Delta \text{SSI}$ and $\Delta \text{Acc}$ across all phenomena and layers, to test whether changes in SSI reflect actual behavioral improvement on syntactic tasks.

\subsection{Ablation Testing}
To validate the functional relevance of supporting neurons (with high SSI), we perform ablation experiments. Specifically, we compare:

\textbf{Targeted ablation}: Zeroing out neurons identified as syntax-sensitive;
    
\textbf{Random ablation}: Zeroing out an equal number of randomly selected neurons.

We then measure the change in PPL on grammatical and ungrammatical sentences. Perplexity reflects how well the model predicts a given sentence---lower $\text{PPL}$ indicates better performance, while higher $\text{PPL}$ indicates increased uncertainty or worse predictions. High-SSI neurons are considered functionally relevant if their ablation leads to a higher increase in perplexity compared to random ablation.

\subsection{Model Training Setup}

To examine the dynamics and variability of syntactic specialization, we trained five instances of the GPT-2 small architecture (124M parameters) under controlled experimental conditions. Three models (GPT-2\textsubscript{Seed1}, GPT-2\textsubscript{Seed7}, and GPT-2\textsubscript{Seed7-Large}) were used for comparative analysis: Two were trained on a small-scale dataset with different random seeds (seed~1 and seed~7), while the third used seed~7 but was trained on a larger-scale corpus. The small-scale dataset consists of 7\,GB of high-quality text, while the large-scale dataset comprises 13\,GB of open-domain content from the latest English Wikipedia dump\footnote{\url{https://dumps.wikimedia.org/enwiki/latest/}}. 

An additional model (GPT-2\textsubscript{Seed123}) was trained on the small-scale dataset using seed~123 to facilitate seed-specific analysis (see \ref{sec:seedsanlaysis}). All models followed the same optimization and training protocol, detailed alongside the exact composition of the small-scale dataset in Appendix~\ref{app:training_config}.

Checkpoint snapshots were saved at multiple training steps for all models, except GPT-2\textsubscript{Seed123}, which was analyzed only at its final state. Each checkpoint is indexed by both (1)~the cumulative number of training data (i.e., at $2^0$, $2^1$, $2^2$, \ldots, $2^{11}$ million tokens)—and (2)~epoch number, allowing fine-grained tracking of syntactic development over time. In total, 46 checkpoints were collected across all models. All code and checkpoints will be released upon publication to support future research.

\section{Experiments}
We evaluate syntactic specialization in LLMs using the proposed SSI. Our experiments are structured to first validate SSI as a meaningful and functionally relevant metric, and then apply it to study the developmental dynamics, variability, and architectural influences on syntactic specialization.

\subsection{Validating the Syntactic Sensitivity Index}
\subsubsection{Correlation with Syntactic Task Performance}
To assess whether SSI reflects actual syntactic competence, we examined its correlation with grammaticality judgment accuracy during training for both GPT-2 \citep{Radford2019} and Pythia-160M \citep{Van2025} models (checkpoints trained on between 0 and 2048 million tokens). For each checkpoint, we computed the change in SSI relative to the final model ($\Delta$SSI) and compared it to the corresponding change in syntactic accuracy ($\Delta$Accuracy) on BLiMP-style classification tasks. A linear mixed-effects model revealed a significant relationship between scaled $\Delta$SSI and $\Delta$Accuracy (for GPT-2\textsubscript{Seed1} and GPT-2\textsubscript{Seed7}, $\beta = 0.15$, $t = 9.50$, $p < .001$; for Pythia, $\beta = 0.14$, $t = 6.37$, $p < .001$), indicating that increases in syntactic selectivity are reliably associated with improvements in syntactic performance. This relationship held while accounting for random variation across layers, seeds, and model families, supporting the interpretation of SSI as a meaningful indicator of emergent syntactic competence during training.

To further assess the developmental sensitivity of SSI, we compared it against traditional probing methods that train supervised classifiers (e.g., SVMs or regressors) to predict syntactic labels from hidden representations \cite{kissane2025probing}. While such methods can capture correlational structure, they often rely on superficial cues or dataset-specific heuristics. We computed the change from the untrained model ($0$M tokens) to the fully trained model ($2048$M tokens) for each metric. SSI showed a significant increase (mean $\Delta = 0.0208$, $t = 12.94$, $p < .001$), indicating a clear growth in structured internal representations. In contrast, both SVM ($Mean = -0.0072$, $t = -0.70$, $p = 0.482$) and regression probes ($Mean = -0.0076$, $t = -0.72$, $p = 0.470$) showed no significant improvement. These results suggest that SSI better captures meaningful structural development during training, whereas supervised probes may remain insensitive to internal specialization, particularly in early learning stages (for details, see Appendix~\ref{sec:appendix.1}).

\subsubsection{Neuron-Level Specialization and Functional Validation}
We extend our analysis to the neuron level by identifying activation dimensions with consistently high SSI scores across linguistic phenomena. For GPT-2, the selected neurons averaged $467$ ($SD = 15$), corresponding to the top $5\%$ of $9{,}216$ total neurons. For Pythia-2.8B, an average of $2{,}890$ neurons ($SD = 1{,}052$) were selected, comprising approximately $3.53\%$ of the full neuron set ($817{,}920$ neurons).
\begin{figure}[h]
\centering
\includegraphics[width=\columnwidth]{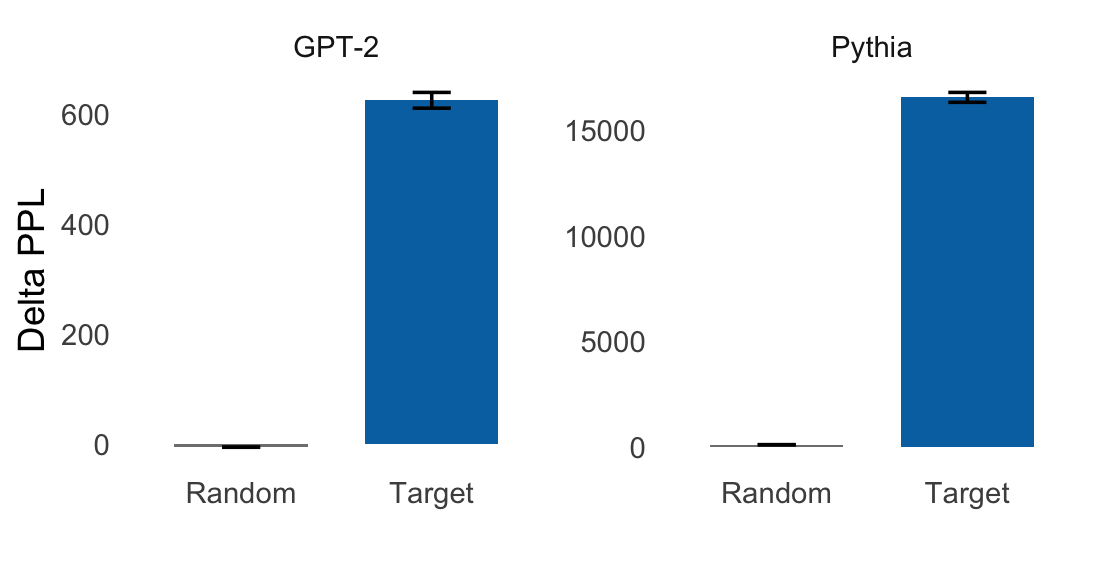}
\caption{ The neuron ablation result with 3 SD of original GPT-2 (left) and Pythia 2.8B (right).}
\label{fig:image2}
\end{figure}

Ablating high-SSI neurons resulted in significantly greater increases in perplexity on grammatical sentences than randomly ablating an equal number of neurons, in both GPT-2 and Pythia (see Figure~\ref{fig:image2}). For GPT-2, ablation of high-SSI neurons increased perplexity by an average of $631$ points, compared to a negligible effect from random ablation ($t = 43.92$, $df = 66{,}999$, $p < .001$). A similar pattern was observed in Pythia, with a mean difference of $16{,}414$ points ($t = 70.22$, $df = 66{,}999$, $p < .001$). These results confirm that high-SSI neurons are not only structurally selective but also functionally necessary for accurate syntactic predictions.

\begin{figure}[htbp]
\centering
\includegraphics[width=\columnwidth]{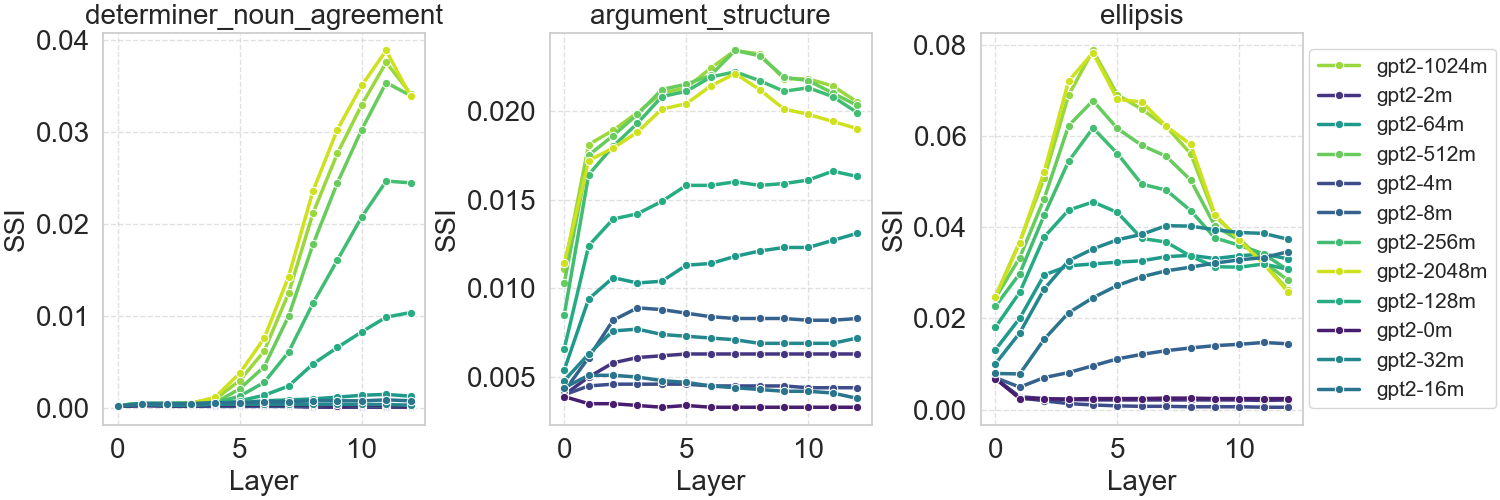}

\caption{The developmental dynamics analysis. Each line represents a model checkpoint (from 0 to 2048M tokens, color-coded from dark blue to yellow), showing the SSI values across layers for a given training stage.}
\label{fig:image3}
\end{figure}

\subsection{Developmental Dynamics of Syntactic Specialization}
We next investigate how syntactic specialization emerges over the course of training. Transformer models, though initialized with random weights and structurally uniform layers, do not encode syntactic knowledge by default. Instead, we observe that syntactic specialization—as measured by the SSI—\textbf{develops gradually}, reflecting a process of cumulative abstraction driven by data exposure and learning dynamics (see Figure~\ref{fig:image3}).

Early in training ($\leq$2M tokens), SSI remains near zero, indicating minimal differentiation across layers or phenomena. As training progresses, distinct developmental trajectories emerge, with SSI increasing steadily across checkpoints. Notably, different linguistic phenomena exhibit \textbf{differential acquisition curves}: Determiner--noun agreement  (e.g., this book vs. *these book) emerges more slowly and peaks in upper layers, indicating that hierarchical agreement relations are abstracted later and at greater depth. \emph{Argument structure} (e.g., That guy remembers vs. *That guy looked like), in contrast, specializes in \textbf{mid layers}, showing a steep rise early on, possibly reflecting the need for deeper compositional understanding. Ellipsis (e.g., She can play the violin, and he can too) shows the earliest signs of differentiation and tends to specialize in lower layers, suggesting reliance on shallow or surface-level cues (see Appendix~\ref{sec:appendix.2} for details).

The developmental curve reveals that syntactic specialization is \textbf{not baked in by initialization}, but instead \textbf{emerges gradually through learning}. Also, it indicated that different acquisition timelines for different syntactic constructs \citep{Evanson2023LanguageAD}.

\begin{figure}[h]
\centering
\includegraphics[width=\columnwidth]{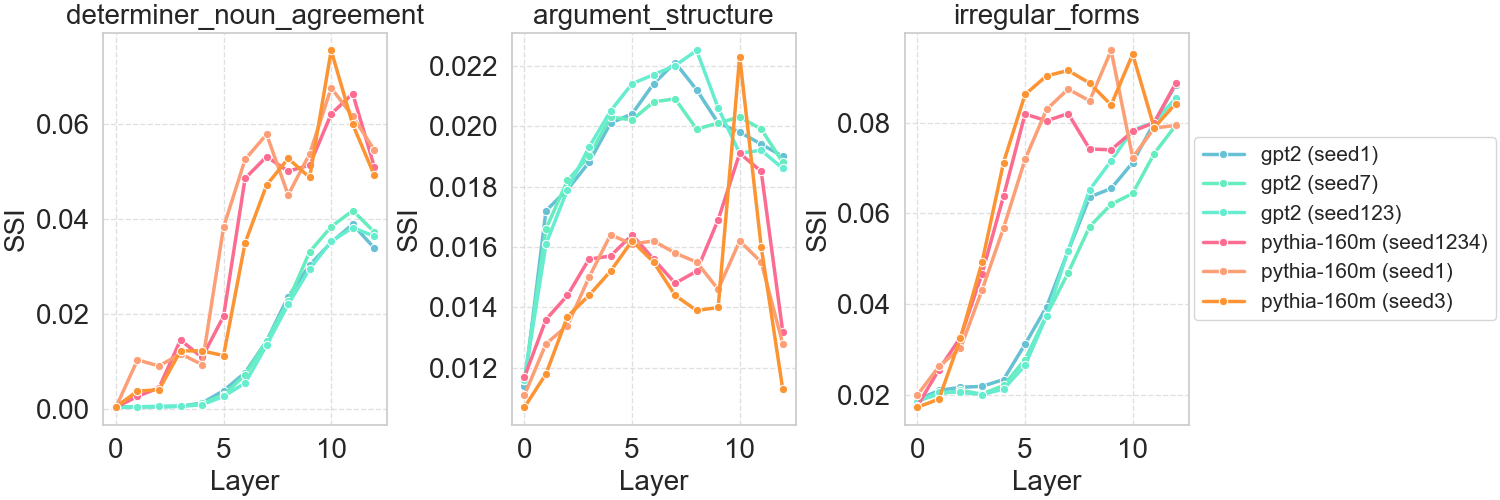}

\caption{Initialization Analysis. Each line represents a model initialized with a different seed. Models from the same family are shown in similar colours.} 

\label{fig:image4}
\end{figure}

\subsection{Initialization-Agnostic Specialization and Architectures Effect}
\label{sec:seedsanlaysis}
To examine the effects of random initialization on syntactic specialization, we trained multiple instances of GPT-2 (124M parameters) using different seeds (1, 7, and 123), and additionally analyzed publicly available Pythia models (160M parameters) initialized with different seeds (1, 3, and 1234). We analyzed layer-wise SSI values averaged across syntactic phenomena (see Figure~\ref{fig:image4}). Pairwise correlations of the averaged SSI across phenomena show that intra-GPT-2 correlations are significantly higher than inter-family (GPT-2 vs. Pythia) correlations (mean~=~0.98 vs.~0.66, $t$~=~22.84, $p$~<~.001), indicating strong consistency within architectures despite different initializations, but notable divergence between families (see Appendix~\ref{sec:appendix.3} for details).

At the \textbf{layer level}, SSI trajectories are highly consistent across seeds, suggesting robust learning of syntax-sensitive representations. However, at the neuron level, we observe \textbf{minimal overlap in top-SSI neuron identities (Mean = $1.49\%$, SD = $1.98\%$)} between seeds (GPT-2\textsubscript{Seed1} and GPT-2\textsubscript{Seed7})---despite similar aggregate SSI scores.

These findings indicate that while models converge on similar syntactic competence, they implement it via \textbf{distinct internal circuits}, reinforcing the idea of \textbf{distributed and non-deterministic representation}.

\subsection{The Critical Period of Specialization}
\label{criticalperiod}
\begin{figure}[h]
\centering
\includegraphics[width=\columnwidth]{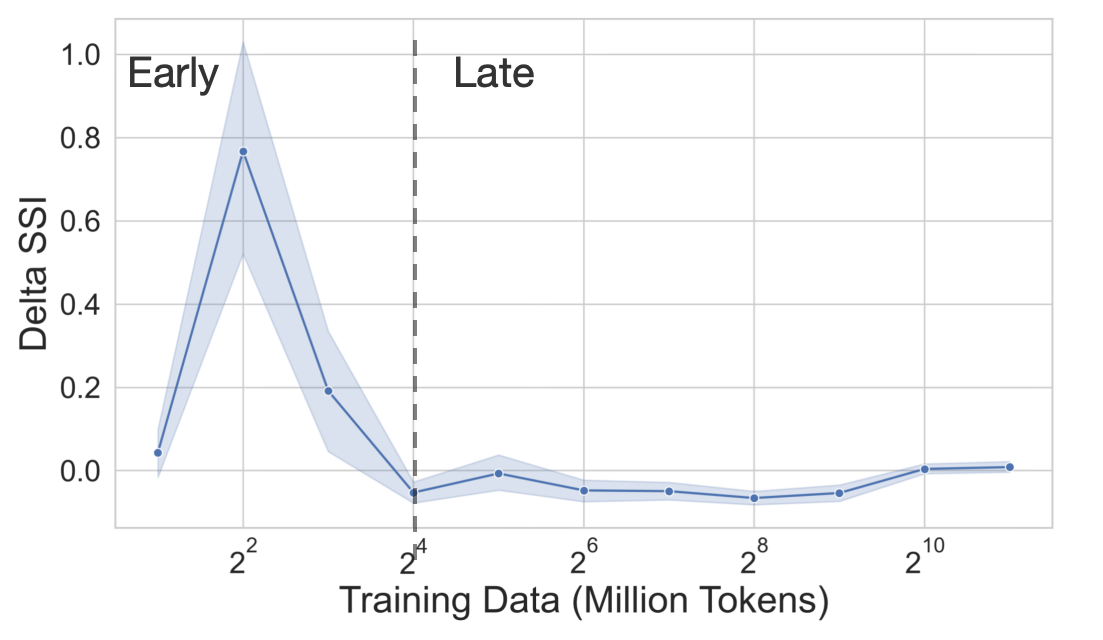}
\caption{Normalized SSI divergence ($\Delta$SSI) between GPT-2\textsubscript{Seed1} and GPT-2\textsubscript{Seed7} across training. Shaded regions indicate 95\% CI.}
\label{fig:image5}
\end{figure}

To investigate the dynamics of syntactic specialization, we analyzed differences in SSI between GPT-2 models initialized with different random seeds (seed~1 and seed~7). Despite identical architectures, models initialized with different seeds show divergent specialization trajectories within the early training phase ($2\text{M}$--$16\text{M}$ tokens), which later converge as training go on, \textbf{suggesting a critical period for syntactic specialization} (see Figure~\ref{fig:image5}).

A mixed-effects model confirmed greater SSI divergence in the late vs.\ early phase ($0.19$ vs.\ $-0.02$, $\beta = 0.229$, $t = 8.77$, $p < .001$; see Appendix~\ref{sec:appendix.4}). In the initial stages of training, the SSI differences between seeds were relatively large, indicating divergent representational development across initializations. However, as training progressed, these differences diminished and eventually converged, reflecting both plasticity in early learning and constraint imposed by the training signal over time.

These findings suggest that the formation of syntax-selective representations is sensitive to early training dynamics but becomes increasingly stable over time. The early phase may serve as a window of heightened architectural malleability, during which inductive biases introduced by initialization can meaningfully shape specialization trajectories---echoing the concept of a critical period in language acquisition observed in human development.

\subsection{Model Scale and Training Data}
\begin{figure}[htbp]
\centering
\includegraphics[width=\columnwidth]{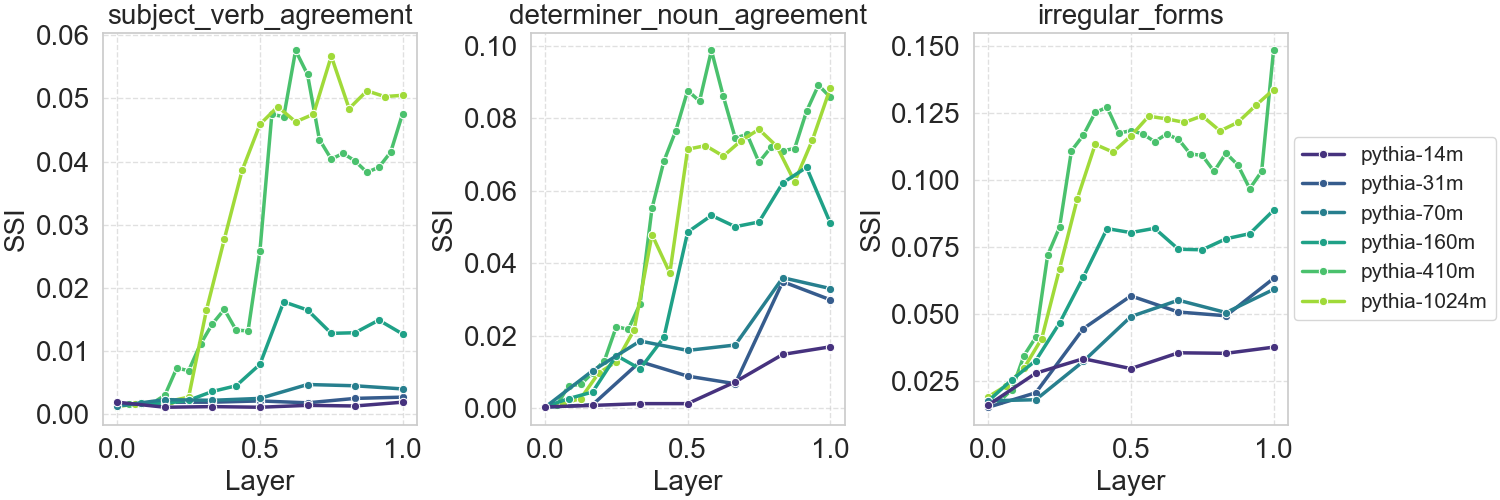}
\caption{Model scale analysis. Each line represents a model of a certain parameter size in the Pythia series. Layer is normalized for comparison.}
\label{fig:image6}
\end{figure}

We examine how model size influences syntactic specialization. In the Pythia series \cite{biderman2023pythiasuiteanalyzinglarge}, SSI increases monotonically with model size ($70\text{M}$ to $1.4\text{B}$ parameters), particularly in phenomena where smaller models show flat SSI distributions, such as subject-verb agreement (see Figure~\ref{fig:image6}). Larger models exhibit more layer-localized specialization, with sharp peaks in specific middle-to-upper layers. Overall specialization patterns are preserved across sizes, with larger models demonstrating greater abstraction and stability (see Figure~\ref{fig:image10} for more linguistic phenomena).
These results confirm that scaling amplifies syntactic abstraction, and that SSI provides a scalable metric for analyzing such effects across architectures.

\begin{figure}[htbp]
    \centering
    \includegraphics[width=\columnwidth]{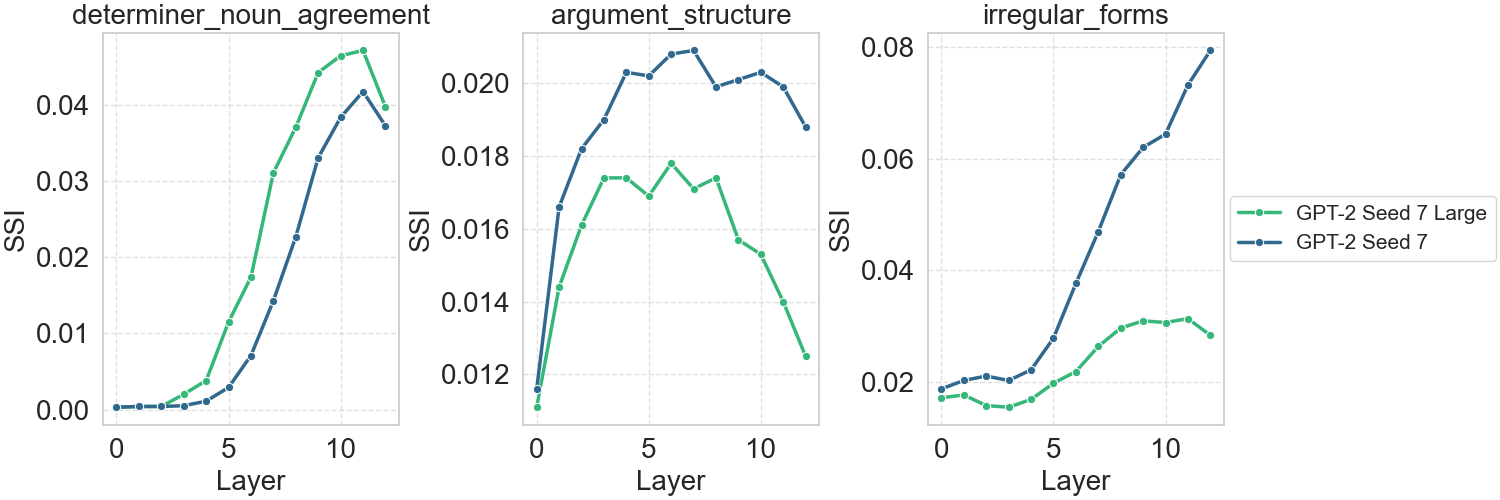}
    \caption{Training data analysis. SSI layer-wise trajectories are shown for three syntactic phenomena. Each line represents a different training condition.}
    \label{fig:image7}
\end{figure}

Beyond scale, by comparing models with identical architecture but different training corpora (GPT-2\textsubscript{Seed7} vs.\ GPT-2\textsubscript{Seed7-Large}), we observed diverging trends depending on the linguistic phenomenon (see Figure~\ref{fig:image7}). For instance, in determiner--noun agreement, the model trained on a larger dataset exhibits higher SSI across layers, potentially reflecting enhanced sensitivity to this rule-based dependency given greater data exposure. In contrast, for irregular forms (e.g., The mushroom goed/went bad), SSI values are lower in the model trained on larger dataset, possibly indicating a more distributed or less sharply localized encoding. One interpretation is that increased training data may support stronger abstraction in more systematic syntactic patterns, while yielding more diffuse representations in cases involving lexical irregularity. However, this pattern is not uniform, and lower SSI does not necessarily imply weaker syntactic encoding. Rather, it may reflect a different representational strategy, particularly for phenomena that are less frequent or less governed by regular grammatical rules (see Appendix~\ref{sec:appendix.5} for details).

Overall, these findings underscore that both model capacity and training data interact to shape how syntax is represented. Larger models and richer datasets tend to enhance abstraction, but the precise nature of specialization depends on the grammatical structure in question. Further investigation is required to disentangle the roles of frequency, complexity, and regularity in shaping these dynamics.

\section{Discussion}

\subsection{Toward a Mechanistic Understanding of Syntactic Specialization}
This work moves beyond surface-level observations of syntactic competence in LLMs by offering a principled, mechanistic account of how syntactic knowledge is acquired, represented, and localized. While prior work has demonstrated that language models can make syntactically informed predictions \cite{hu2024language, qiu2025grammaticality}, our goal was to understand how such structural competence emerges—when during training it appears, where it is represented within the model, and which factors shape its development. By introducing the SSI, we provide a tool that captures the internal differentiation between grammatical and ungrammatical representations across a wide range of syntactic phenomena—\emph{without relying on external supervision or classifiers}. This internalism is crucial: SSI does not require behavioral probes or finetuning, allowing us to observe syntactic structure as it is natively encoded in the geometry of model activations.

One of the most important findings is the tight coupling between SSI and syntactic accuracy: As the model becomes more syntactically competent behaviorally, its internal representations become more structurally differentiated. This strongly supports the validity of SSI as a proxy for syntactic development. Just as children’s grammatical knowledge becomes more fine-grained and stable over time, so too does the model’s internal capacity to separate structural contrasts.

Furthermore, the emergence of SSI follows a clear developmental trajectory—flat early on, rising sharply in mid-training, and plateauing later. This suggests that syntactic specialization is not innate to the architecture, but is gradually constructed through exposure. The identification of this \textbf{"critical period"} of syntactic specialization aligns with theories of staged linguistic development in humans and may inform training curricula or interventions for more efficient structure learning.

\subsection{Scaling, Abstraction, and Emergent Syntax}
We also find that model scale plays a central role in shaping syntactic specialization. Larger models not only exhibit stronger SSI scores, but also show more \textbf{layer-localized abstraction} of syntactic distinctions. This aligns with the intuition that greater capacity enables models to form deeper, more abstract structural generalizations—consistent with scaling laws in language modeling. Yet the scaling is not uniform across phenomena: Certain syntactic categories, like subject-verb agreement, benefit more from increased capacity than others. This suggests that different components of syntactic specialization may emerge along distinct representational pathways.

\section{Conclusion}
Our study provides the first systematic, mechanistic account of how syntactic specialization emerges in transformer language models. By introducing SSI and applying it across training trajectories, seeds, and model scales, we offer an integrated picture of when, where, and how syntax lives in neural models. We hope this work contributes both methodological tools and conceptual clarity to the growing effort to reverse-engineer the internal dynamics of large language models.

\section{Limitations}
While SSI reveals important patterns of syntactic specialization, several limitations remain. First, our analysis relies on sentence embeddings (mean pooling), which may miss the token-level nuance or positional dynamics. Future work could extend SSI to token-level activations or attention maps. Second, while BLiMP provides high-coverage diagnostics, it focuses on binary acceptability—leaving gradient judgments, semantic-pragmatic interactions, and multi-sentence structures underexplored.

Finally, our results raise open theoretical questions: What is the minimal inductive bias or data regime required for syntactic specialization to emerge? Can architectural interventions (e.g., sparsity, recurrence) accelerate or refine this process? And how does syntactic development relate to semantic grounding or world knowledge in LLMs?

\bibliography{custom}

\begin{thebibliography}{39}
\providecommand{\natexlab}[1]{#1}

\bibitem[{AlKhamissi et~al.(2025)AlKhamissi, Tuckute, Tang, Binhuraib, Bosselut, and Schrimpf}]{alkhamissi2025language}
Badr AlKhamissi, Greta Tuckute, Yingtian Tang, Taha Binhuraib, Antoine Bosselut, and Martin Schrimpf. 2025.
\newblock From language to cognition: How llms outgrow the human language network.
\newblock \emph{arXiv preprint arXiv:2503.01830}.

\bibitem[{Ameisen et~al.(2025)Ameisen, Lindsey, Pearce, Gurnee, Turner, Chen, Citro, Abrahams, Carter, Hosmer, Marcus, Sklar, Templeton, Bricken, McDougall, Cunningham, Henighan, Jermyn, Jones, Persic, Qi, Ben~Thompson, Zimmerman, Rivoire, Conerly, Olah, and Batson}]{ameisen2025circuit}
Emmanuel Ameisen, Jack Lindsey, Adam Pearce, Wes Gurnee, Nicholas~L. Turner, Brian Chen, Craig Citro, David Abrahams, Shan Carter, Basil Hosmer, Jonathan Marcus, Michael Sklar, Adly Templeton, Trenton Bricken, Callum McDougall, Hoagy Cunningham, Thomas Henighan, Adam Jermyn, Andy Jones, and 8 others. 2025.
\newblock \href {https://transformer-circuits.pub/2025/attribution-graphs/methods.html} {Circuit tracing: Revealing computational graphs in language models}.
\newblock \emph{Transformer Circuits Thread}.

\bibitem[{Bates et~al.(2015)Bates, M{\"a}chler, Bolker, and Walker}]{bates2015fitting}
Douglas Bates, Martin M{\"a}chler, Ben Bolker, and Steve Walker. 2015.
\newblock Fitting linear mixed-effects models using lme4.
\newblock \emph{Journal of statistical software}, 67:1--48.

\bibitem[{Biderman et~al.(2023)Biderman, Schoelkopf, Anthony, Bradley, O'Brien, Hallahan, Khan, Purohit, Prashanth, Raff, Skowron, Sutawika, and van~der Wal}]{biderman2023pythiasuiteanalyzinglarge}
Stella Biderman, Hailey Schoelkopf, Quentin Anthony, Herbie Bradley, Kyle O'Brien, Eric Hallahan, Mohammad~Aflah Khan, Shivanshu Purohit, USVSN~Sai Prashanth, Edward Raff, Aviya Skowron, Lintang Sutawika, and Oskar van~der Wal. 2023.
\newblock \href {https://arxiv.org/abs/2304.01373} {Pythia: A suite for analyzing large language models across training and scaling}.
\newblock \emph{Preprint}, arXiv:2304.01373.

\bibitem[{Brinkmann et~al.(2025)Brinkmann, Wendler, Bartelt, and Mueller}]{brinkmann2025large}
Jannik Brinkmann, Chris Wendler, Christian Bartelt, and Aaron Mueller. 2025.
\newblock Large language models share representations of latent grammatical concepts across typologically diverse languages.
\newblock \emph{arXiv preprint arXiv:2501.06346}.

\bibitem[{Chen et~al.(2024)Chen, Shwartz-Ziv, Cho, Leavitt, and Saphra}]{chen2024sudden}
Angelica Chen, Ravid Shwartz-Ziv, Kyunghyun Cho, Matthew~L Leavitt, and Naomi Saphra. 2024.
\newblock \href {https://openreview.net/forum?id=MO5PiKHELW} {Sudden drops in the loss: Syntax acquisition, phase transitions, and simplicity bias in {MLM}s}.
\newblock In \emph{The Twelfth International Conference on Learning Representations}.

\bibitem[{Clark et~al.(2019)Clark, Khandelwal, Levy, and Manning}]{clark-etal-2019-bert}
Kevin Clark, Urvashi Khandelwal, Omer Levy, and Christopher~D. Manning. 2019.
\newblock \href {https://doi.org/10.18653/v1/W19-4828} {What does {BERT} look at? an analysis of {BERT}`s attention}.
\newblock In \emph{Proceedings of the 2019 ACL Workshop BlackboxNLP: Analyzing and Interpreting Neural Networks for NLP}, pages 276--286, Florence, Italy. Association for Computational Linguistics.

\bibitem[{Curtiss(1977)}]{curtiss1977}
Susan Curtiss. 1977.
\newblock \emph{Genie: A Psycholinguistic Study of a Modern-Day Wild Child}.
\newblock Academic Press, New York.

\bibitem[{Duan et~al.(2025)Duan, Zhou, Xiao, and Cai}]{duan2025unveiling}
Xufeng Duan, Xinyu Zhou, Bei Xiao, and Zhenguang Cai. 2025.
\newblock Unveiling language competence neurons: A psycholinguistic approach to model interpretability.
\newblock In \emph{Proceedings of the 31st International Conference on Computational Linguistics}, pages 10148--10157.

\bibitem[{Fedorenko and Kanwisher(2009)}]{fedorenko2009}
Evelina Fedorenko and Nancy Kanwisher. 2009.
\newblock Neuroimaging of language: why hasn't a clearer picture emerged?
\newblock \emph{Language and Linguistics Compass}, 3(4):839--865.

\bibitem[{Fedorenko et~al.(2024)Fedorenko, Piantadosi, and Gibson}]{fedorenko2024language}
Evelina Fedorenko, Steven~T Piantadosi, and Edward~AF Gibson. 2024.
\newblock Language is primarily a tool for communication rather than thought.
\newblock \emph{Nature}, 630(8017):575--586.

\bibitem[{Friederici(2018)}]{friederici2018neural}
Angela~D Friederici. 2018.
\newblock The neural basis for human syntax: Broca's area and beyond.
\newblock \emph{Current opinion in behavioral sciences}, 21:88--92.

\bibitem[{Hall et~al.(2018)Hall, Li, and Dye}]{hall2018}
Wyatte~C. Hall, Dongmei Li, and Tim D.~V. Dye. 2018.
\newblock \href {https://doi.org/10.2105/AJPH.2018.304498} {Influence of hearing loss on child behavioral and home experiences}.
\newblock \emph{American Journal of Public Health}, 108(8):1079--1081.
\newblock Epub 2018 Jun 21.

\bibitem[{Hertz-Pannier et~al.(2002)Hertz-Pannier, Chiron, Jambaque, Renaux-Kieffer, Van~de Moortele, Delalande, Fohlen, Brunelle, and Le~Bihan}]{hertz2002}
L.~Hertz-Pannier, C.~Chiron, I.~Jambaque, V.~Renaux-Kieffer, P.~F. Van~de Moortele, O.~Delalande, M.~Fohlen, F.~Brunelle, and D.~Le~Bihan. 2002.
\newblock Late plasticity for language in a child's non-dominant hemisphere: a pre- and post-surgery fmri study.
\newblock \emph{Brain}, 125(2):361--372.

\bibitem[{Hu et~al.(2024)Hu, Mahowald, Lupyan, Ivanova, and Levy}]{hu2024language}
Jennifer Hu, Kyle Mahowald, Gary Lupyan, Anna Ivanova, and Roger Levy. 2024.
\newblock Language models align with human judgments on key grammatical constructions.
\newblock \emph{Proceedings of the National Academy of Sciences}, 121(36):e2400917121.

\bibitem[{Johnson and Newport(1989)}]{johnson1989}
Jacqueline~S. Johnson and Elissa~L. Newport. 1989.
\newblock Critical period effects in second language learning: The influence of maturational state on the acquisition of english as a second language.
\newblock \emph{Cognitive Psychology}, 21(1):60--99.

\bibitem[{Keshavan et~al.(2014)Keshavan, Giedd, Lau, Lewis, and Paus}]{Keshavan2014}
MS~Keshavan, J~Giedd, JY~Lau, DA~Lewis, and T~Paus. 2014.
\newblock \href {https://doi.org/10.1016/S2215-0366(14)00081-9} {Changes in the adolescent brain and the pathophysiology of psychotic disorders}.
\newblock \emph{Lancet Psychiatry}, 1(7):549--558.

\bibitem[{Kissane et~al.(2025)Kissane, Schilling, and Krauss}]{kissane2025probing}
Hassane Kissane, Achim Schilling, and Patrick Krauss. 2025.
\newblock Probing internal representations of multi-word verbs in large language models.
\newblock \emph{arXiv preprint arXiv:2502.04789}.

\bibitem[{Klein et~al.(2023)Klein, Berger, Goucha, Friederici, and Grosse~Wiesmann}]{klein2023children}
Cheslie~C Klein, Philipp Berger, Tom{\'a}s Goucha, Angela~D Friederici, and Charlotte Grosse~Wiesmann. 2023.
\newblock Children’s syntax is supported by the maturation of ba44 at 4 years, but of the posterior sts at 3 years of age.
\newblock \emph{Cerebral Cortex}, 33(9):5426--5435.

\bibitem[{Kuznetsova et~al.(2017)Kuznetsova, Brockhoff, and Christensen}]{kuznetsova2017lmertest}
Alexandra Kuznetsova, Per~B Brockhoff, and Rune~HB Christensen. 2017.
\newblock lmertest package: tests in linear mixed effects models.
\newblock \emph{Journal of statistical software}, 82:1--26.

\bibitem[{Lakretz et~al.(2019)Lakretz, Kruszewski, Desbordes, Hupkes, Dehaene, and Baroni}]{lakretz2019}
Yair Lakretz, German Kruszewski, Theo Desbordes, Dieuwke Hupkes, Stanislas Dehaene, and Marco Baroni. 2019.
\newblock \href {https://doi.org/10.18653/v1/N19-1002} {{The emergence of number and syntax units in LSTM language models}}.
\newblock In \emph{Proceedings of the 2019 Conference of the North American Chapter of the Association for Computational Linguistics: Human Language Technologies, Volume 1 (Long and Short Papers)}, pages 11--20, Stroudsburg, PA, USA. Association for Computational Linguistics.

\bibitem[{Lenneberg(1967)}]{lenneberg1967}
Eric~H. Lenneberg. 1967.
\newblock \emph{Biological Foundations of Language}.
\newblock Wiley, New York.

\bibitem[{Linzen et~al.(2016)Linzen, Dupoux, and Goldberg}]{linzen-etal-2016-assessing}
Tal Linzen, Emmanuel Dupoux, and Yoav Goldberg. 2016.
\newblock \href {https://doi.org/10.1162/tacl_a_00115} {Assessing the ability of {LSTM}s to learn syntax-sensitive dependencies}.
\newblock \emph{Transactions of the Association for Computational Linguistics}, 4:521--535.

\bibitem[{Liu et~al.(2025)Liu, Gao, Sun, Ge, Liu, Han, and Hu}]{liu2025brain}
Yiheng Liu, Xiaohui Gao, Haiyang Sun, Bao Ge, Tianming Liu, Junwei Han, and Xintao Hu. 2025.
\newblock Brain-inspired exploration of functional networks and key neurons in large language models.
\newblock \emph{arXiv preprint arXiv:2502.20408}.

\bibitem[{Liu et~al.(2024)Liu, Shen, Zhu, Xu, Qian, Song, Zhang, Tang, Zhang, Yang et~al.}]{liu2024zhoblimp}
Yikang Liu, Yeting Shen, Hongao Zhu, Lilong Xu, Zhiheng Qian, Siyuan Song, Kejia Zhang, Jialong Tang, Pei Zhang, Baosong Yang, and 1 others. 2024.
\newblock Zhoblimp: a systematic assessment of language models with linguistic minimal pairs in chinese.
\newblock \emph{arXiv preprint arXiv:2411.06096}.

\bibitem[{Marvin and Linzen(2018)}]{marvin2018targeted}
Rebecca Marvin and Tal Linzen. 2018.
\newblock Targeted syntactic evaluation of language models.
\newblock \emph{arXiv preprint arXiv:1808.09031}.

\bibitem[{Mueller et~al.(2022)Mueller, Xia, and Linzen}]{mueller2022causal}
Aaron Mueller, Yu~Xia, and Tal Linzen. 2022.
\newblock Causal analysis of syntactic agreement neurons in multilingual language models.
\newblock \emph{arXiv preprint arXiv:2210.14328}.

\bibitem[{Olsson et~al.(2022)Olsson, Elhage, Nanda, Joseph, DasSarma, Henighan, Mann, Askell, Bai, Chen et~al.}]{olsson2022context}
Catherine Olsson, Nelson Elhage, Neel Nanda, Nicholas Joseph, Nova DasSarma, Tom Henighan, Ben Mann, Amanda Askell, Yuntao Bai, Anna Chen, and 1 others. 2022.
\newblock In-context learning and induction heads.
\newblock \emph{arXiv preprint arXiv:2209.11895}.

\bibitem[{Pénicaud et~al.(2013)Pénicaud, Klein, Zatorre, Chen, Witcher, Hyde, and Mayberry}]{penicaud2013}
S~Pénicaud, D~Klein, RJ~Zatorre, JK~Chen, P~Witcher, K~Hyde, and RI~Mayberry. 2013.
\newblock \href {https://doi.org/10.1016/j.neuroimage.2012.09.076} {Structural brain changes linked to delayed first language acquisition in congenitally deaf individuals}.
\newblock \emph{Neuroimage}, 66:42--49.

\bibitem[{Qiu et~al.(2025)Qiu, Duan, and Cai}]{qiu2025grammaticality}
Zhuang Qiu, Xufeng Duan, and Zhenguang~G Cai. 2025.
\newblock Grammaticality representation in chatgpt as compared to linguists and laypeople.
\newblock \emph{Humanities and Social Sciences Communications}, 12(1):1--15.

\bibitem[{Radford et~al.(2019)Radford, Wu, Child et~al.}]{Radford2019}
Alec Radford, Jeffrey Wu, Rewon Child, and 1 others. 2019.
\newblock Language models are unsupervised multitask learners.
\newblock \emph{OpenAI Blog}, 1.
\newblock Available at \url{https://openai.com/blog/better-language-models/}.

\bibitem[{Raghu et~al.(2017)Raghu, Gilmer, Yosinski, and Sohl-Dickstein}]{raghu2017svcca}
Maithra Raghu, Justin Gilmer, Jason Yosinski, and Jascha Sohl-Dickstein. 2017.
\newblock Svcca: Singular vector canonical correlation analysis for deep learning dynamics and interpretability.
\newblock \emph{Advances in neural information processing systems}, 30.

\bibitem[{Tang et~al.(2024)Tang, Luo, Huang, Zhang, Wang, Zhao, Wei, and Wen}]{tang2024language}
Tianyi Tang, Wenyang Luo, Haoyang Huang, Dongdong Zhang, Xiaolei Wang, Xin Zhao, Furu Wei, and Ji-Rong Wen. 2024.
\newblock Language-specific neurons: The key to multilingual capabilities in large language models.
\newblock \emph{arXiv preprint arXiv:2402.16438}.

\bibitem[{van~der Wal et~al.(2025)van~der Wal, Lesci, M{\"u}ller-Eberstein, Saphra, Schoelkopf, Zuidema, and Biderman}]{Van2025}
Oskar van~der Wal, Pietro Lesci, Max M{\"u}ller-Eberstein, Naomi Saphra, Hailey Schoelkopf, Willem Zuidema, and Stella Biderman. 2025.
\newblock Polypythias: Stability and outliers across fifty language model pre-training runs.
\newblock In \emph{{The Thirteenth International Conference on Learning Representations}}.

\bibitem[{Vicari et~al.(2000)Vicari, Albertoni, Chilosi, Cipriani, Cioni, and Bates}]{vicari2000}
S.~Vicari, A.~Albertoni, A.~M. Chilosi, P.~Cipriani, G.~Cioni, and E.~Bates. 2000.
\newblock Plasticity and reorganization during language development in children with early brain injury.
\newblock \emph{Cortex}, 36(1):31--46.

\bibitem[{Voita et~al.(2019)Voita, Talbot, Moiseev, Sennrich, and Titov}]{voita-etal-2019-analyzing}
Elena Voita, David Talbot, Fedor Moiseev, Rico Sennrich, and Ivan Titov. 2019.
\newblock \href {https://doi.org/10.18653/v1/P19-1580} {Analyzing multi-head self-attention: Specialized heads do the heavy lifting, the rest can be pruned}.
\newblock In \emph{Proceedings of the 57th Annual Meeting of the Association for Computational Linguistics}, pages 5797--5808, Florence, Italy. Association for Computational Linguistics.

\bibitem[{Wang et~al.(2022)Wang, Variengien, Conmy, Shlegeris, and Steinhardt}]{wang2022interpretability}
Kevin Wang, Alexandre Variengien, Arthur Conmy, Buck Shlegeris, and Jacob Steinhardt. 2022.
\newblock Interpretability in the wild: a circuit for indirect object identification in gpt-2 small.
\newblock \emph{arXiv preprint arXiv:2211.00593}.

\bibitem[{Warstadt et~al.(2020)Warstadt, Parrish, Liu, Mohananey, Peng, Wang, and Bowman}]{warstadt-etal-2020-blimp-benchmark}
Alex Warstadt, Alicia Parrish, Haokun Liu, Anhad Mohananey, Wei Peng, Sheng-Fu Wang, and Samuel~R. Bowman. 2020.
\newblock \href {https://doi.org/10.1162/tacl_a_00321} {{BL}i{MP}: The benchmark of linguistic minimal pairs for {E}nglish}.
\newblock \emph{Transactions of the Association for Computational Linguistics}, 8:377--392.

\bibitem[{Zeng et~al.(2025)Zeng, Han, Chen, and Yu}]{zeng-etal-2025-converging}
Hongchuan Zeng, Senyu Han, Lu~Chen, and Kai Yu. 2025.
\newblock \href {https://aclanthology.org/2025.coling-main.707/} {Converging to a lingua franca: Evolution of linguistic regions and semantics alignment in multilingual large language models}.
\newblock In \emph{Proceedings of the 31st International Conference on Computational Linguistics}, pages 10602--10617, Abu Dhabi, UAE. Association for Computational Linguistics.

\end{thebibliography}
\newpage

\appendix

\section{Appendix}
\label{sec:appendix}

\subsection{Statistical Analysis of the Relationship Between SSI and Syntactic Accuracy}
\label{sec:appendix.1}
To evaluate whether the SSI reflects the emergence of syntactic competence during training, we conducted both correlational and mixed-effects modeling analyses using data from GPT-2 (124M; seeds~1 and~7) and Pythia-160M. Each model was evaluated at 12 training checkpoints: $0$, $2$, $4$, $8$, $16$, $32$, $64$, $128$, $256$, $512$, $1024$, and $2048$ million tokens. At each checkpoint, we computed $\Delta$SSI and $\Delta$Accuracy at the layer--phenomenon level, defined as the absolute difference from the final ($2048$M-token) model. To ensure comparability across model families and training scales, these values were standardized ($z$-scored) within each model group prior to statistical analysis:

\[
\text{z$\Delta$SSI} = \frac{|\Delta \text{SSI}| - \mu_{\Delta \text{SSI}}}{\sigma_{\Delta \text{SSI}}}
\]

\[
\text{z$\Delta$Accuracy} = \frac{|\Delta \text{Accuracy}| - \mu_{\Delta \text{Accuracy}}}{\sigma_{\Delta \text{Accuracy}}}
\]

\textbf{Mixed-Effects Modeling} We fit a series of linear mixed-effects models predicting standardized $\Delta$Accuracy from standardized $\Delta$SSI using the \texttt{lme4} \cite{bates2015fitting} and \texttt{lmerTest} \cite{kuznetsova2017lmertest} packages in R. For GPT-2 (combining seeds~1 and~7), the best-fitting model included fixed effects of $\Delta$SSI and random intercepts for Layer, Seed and phenomenon:

\begin{multline*}
\text{z$\Delta$Accuracy} \sim \text{z$\Delta$SSI} + (1|\text{Layer}) +\\
(1|\text{Seed})+ (1|\text{Phenomenon})
\end{multline*}

For Pythia-160M, we fit a comparable model with random intercepts for Layer and Phenomenon.

\begin{multline*}
\text{z$\Delta$Accuracy} \sim \text{z$\Delta$SSI} + (1|\text{Layer})+ \\
(1|\text{Phenomenon})
\end{multline*}

Likelihood ratio tests ($\alpha = 0.2$) showed that including random slopes for $\Delta$SSI by Layer, Seed, and Phenomenon significantly improved model fit for GPT-2; for Pythia, only the random slope by Phenomenon was retained.

These results confirm that increases in SSI are closely aligned with gains in syntactic accuracy, especially in the transition from untrained to fully trained models. Notably, the stronger correlation observed in Pythia may reflect architectural or data-related differences in syntactic learning dynamics. Together, the mixed-effects and correlational results converge to support the validity of SSI as a meaningful and scalable indicator of emerging syntactic competence in neural language models.

\begin{figure*}[h]
\centering
\includegraphics[width=\textwidth]{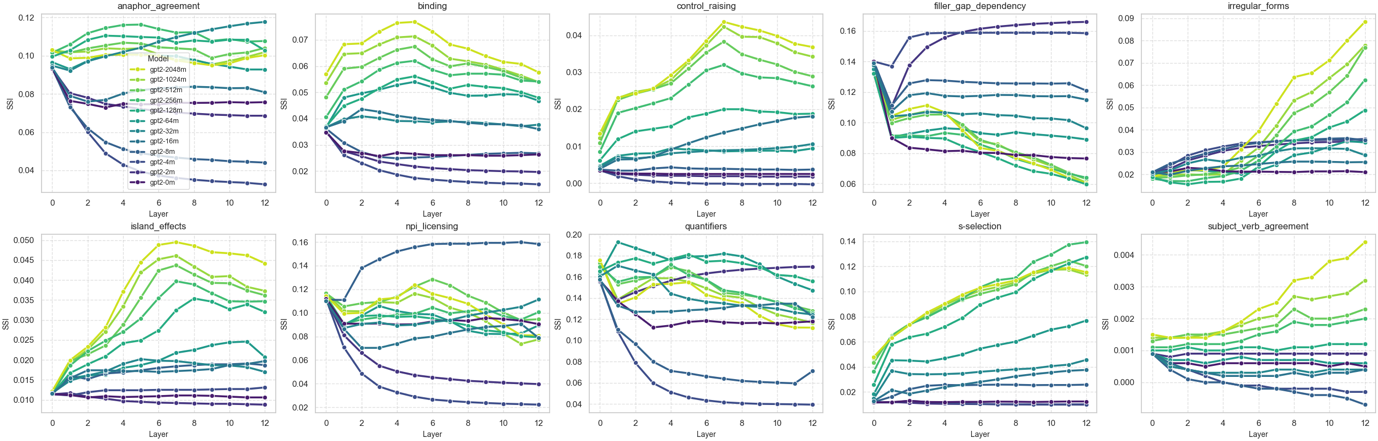}
\caption{ Developmental trajectories of SSI across syntactic phenomena in GPT-2 (seed 1 and seed 7). }
\label{fig:image8}
\end{figure*}

\subsection{Developmental Analysis of SSI During Training}
\label{sec:appendix.2}

Each subplot shows SSI values across layers at multiple training checkpoints (from $0$ to $2048$\text{M} tokens, color-coded from dark blue to yellow) for a distinct syntactic phenomenon. The progression illustrates that syntactic selectivity emerges gradually during training, with distinct acquisition timelines and layer profiles across phenomena.

To investigate the emergence of syntactic selectivity over training, we fit a series of linear mixed-effects models. The dependent variable was SSI, with training data quantity (in tokens) as the primary predictor. The baseline model included training data as a fixed effect and random intercepts for Layer, Seed, and Phenomenon:

\begin{multline*}
\text{SSI} \sim \text{Training\_data} + (1|\text{Layer}) + \\
(1|\text{Seed}) + (1|\text{Phenomenon})
\end{multline*}

To assess whether model fit improved with additional complexity, we tested variants that incorporated random slopes for \texttt{Training\_data} by each grouping factor. Likelihood ratio tests ($\alpha = 0.2$) indicated that these more complex models did not significantly improve fit over the baseline, and so the simpler specification was retained.

In the final model, the effect of training data was highly significant ($\beta = 0.006$, SE $= 0.00$, $t = 15.17$, $p < .001$), confirming a systematic increase in SSI with additional training exposure. This trend held across multiple layers, seeds, and syntactic phenomena. The early training phase showed minimal syntactic sensitivity, with SSI emerging gradually and reaching peak selectivity in mid-to-late layers as training progressed. These findings underscore the dynamic, learned nature of syntactic specialization in transformer models.

\subsection{Seed analysis for Initialization parameter and Architecture effect}
\label{sec:appendix.3}
To quantify the similarity of syntactic selectivity across models, we computed pairwise Pearson correlation coefficients between the SSI profiles of all independently trained models. For each model $m$, we define its SSI profile as a vector of layer-wise values averaged across all linguistic phenomena:

\[
\text{SSI}^{(m)} = [s^{(m)}_1, s^{(m)}_2, \dots, s^{(m)}_L]
\]

where $s^{(m)}_\ell$ denotes the SSI at layer $\ell$, and $L$ is the total number of layers.

For each model pair $(i, j)$, Pearson correlation was computed as:

\[
r_{ij} = \frac{\sum_{\ell=1}^{L} (s^{(i)}_\ell - \bar{s}^{(i)})(s^{(j)}_\ell - \bar{s}^{(j)})}{\sqrt{\sum_{\ell=1}^{L} (s^{(i)}_\ell - \bar{s}^{(i)})^2} \sqrt{\sum_{\ell=1}^{L} (s^{(j)}_\ell - \bar{s}^{(j)})^2}}
\]

where $\bar{s}^{(i)}$ and $\bar{s}^{(j)}$ are the mean SSI values across layers for models $i$ and $j$, respectively.

To compare intra-family (within GPT-2) and inter-family (between GPT-2 and Pythia) correlation distributions, we applied a one-sided Welch's $t$-test. The test statistic is given by:

\[
t = \frac{\bar{r}_{\text{intra}} - \bar{r}_{\text{inter}}}{\sqrt{\frac{s^2_{\text{intra}}}{n_{\text{intra}}} + \frac{s^2_{\text{inter}}}{n_{\text{inter}}}}}
\]

with degrees of freedom estimated via:

\[
df = \frac{\left( \frac{s^2_{\text{intra}}}{n_{\text{intra}}} + \frac{s^2_{\text{inter}}}{n_{\text{inter}}} \right)^2}{\frac{(s^2_{\text{intra}})^2}{n_{\text{intra}}^2(n_{\text{intra}} - 1)} + \frac{(s^2_{\text{inter}})^2}{n_{\text{inter}}^2(n_{\text{inter}} - 1)}}
\]

where $\bar{r}$, $s^2$, and $n$ denote the sample mean, variance, and size of the correlation distributions for each group. Self-correlations and duplicate pairings were excluded from all calculations.

\subsection{SSI Divergence Analysis Across Training Phases}
\label{sec:appendix.4}
To quantify the temporal dynamics of seed-induced variability in syntactic specialization, we analyzed $\Delta$SSI between two GPT-2 models trained with different random seeds (seed~1 and seed~7). $\Delta$SSI was computed at each training checkpoint by taking the absolute difference in SSI values across layers and phenomena between the two models. To enable comparison across stages of training and across phenomena, $\Delta$SSI values were normalized by the mean SSI of the two seeds at each data point.

We grouped training checkpoints into early ($\leq$16M tokens) and late ($>$16M tokens) phases and fit a linear mixed-effects model predicting $\Delta$SSI from phase, with random intercepts and slopes for linguistic phenomenon and model layer. The model revealed a significant increase in $\Delta$SSI in the late phase compared to the early phase ($\beta = 0.229$, $t = 8.77$, $p < .001$), indicating greater divergence in specialization trajectories later in training.

\subsection{Effects of Model Scale and Training Data on Syntactic Specialization}
\label{sec:appendix.5}

\begin{figure*}[h]
\centering
\includegraphics[width=\textwidth]{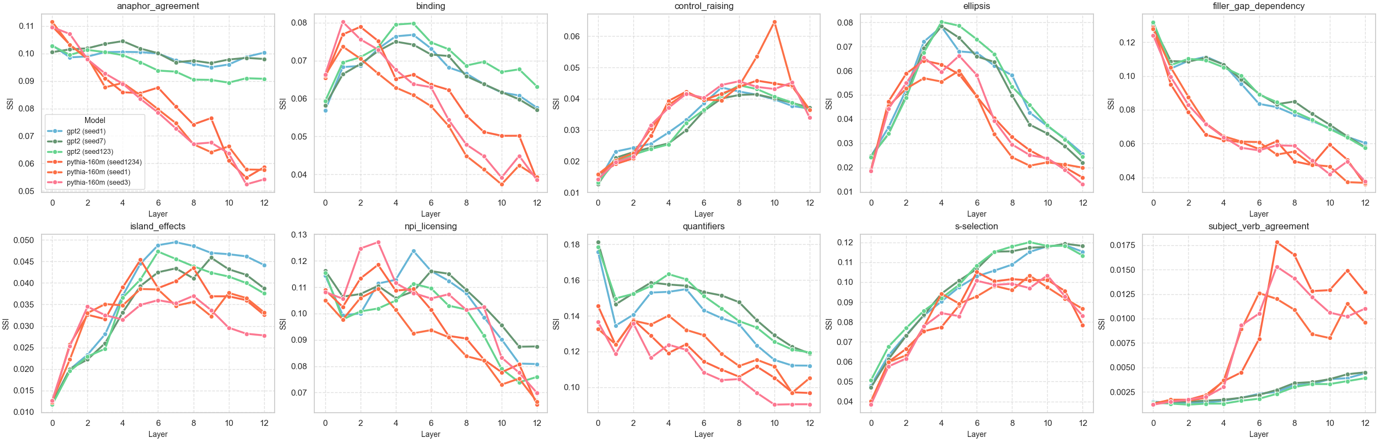}
\caption{SSI trajectories across seeds and architectures for ten syntactic phenomena. Each line represents a model instance with a different initialization seed (GPT-2: green/blue; Pythia: red/orange), showing layer-wise SSI values for a given syntactic phenomenon.}
\label{fig:image9}
\end{figure*}
\begin{figure*}[h]
\centering
\includegraphics[width=\textwidth]{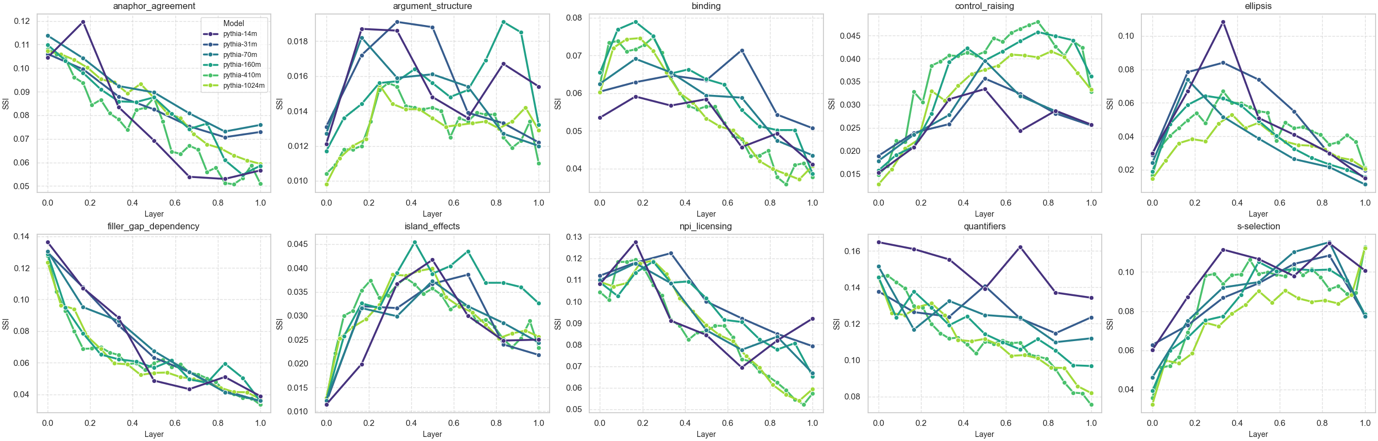}
\caption{Each line represents a model from the Pythia series (70M–1.4B parameters), with SSI plotted across normalized layer depth. }
\label{fig:image10}
\end{figure*}
\begin{figure*}[t]
\vspace*{-40em}
\includegraphics[width=\textwidth]{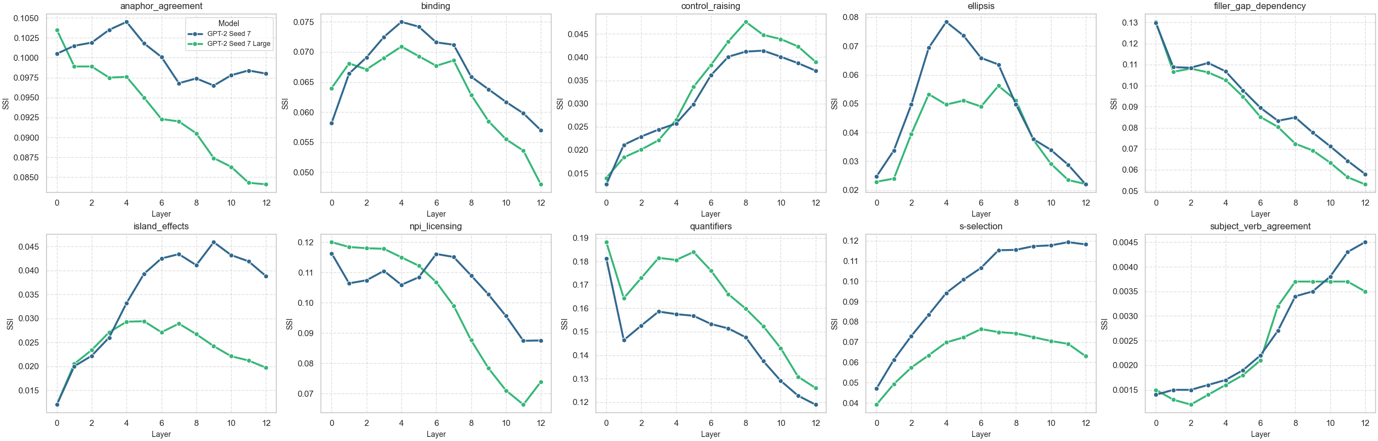}
\caption{Layer-wise SSI trajectories are shown for multiple syntactic phenomena in GPT-2 models trained on small (7GB) vs. large (13GB) datasets. Each line corresponds to a model/seed pairing. Effects of training data scale vary by phenomenon, reflecting a complex interaction between data quantity and the nature of syntactic structures being learned.}
\label{fig:image11}
\end{figure*}

\subsection{Training Configuration}
\label{app:training_config}

All models were trained using the same optimization settings to ensure fair comparisons across experimental conditions. Training was conducted with the following hyperparameters:

\begin{itemize}
  \item \textbf{Learning rate:} $1\times10^{-4}$
  \item \textbf{Optimizer:} Adam
  \item \textbf{Gradient accumulation steps:} 8
  \item \textbf{Per-device batch size:} 20
  \item \textbf{Mixed-precision training:} Enabled (\texttt{fp16})
  \item \textbf{Training hardware:} Eight NVIDIA GeForce RTX 3090 GPUs using data-parallelism
\end{itemize}

The small-scale dataset included the English part of following resources:

\begin{itemize}
\item \textbf{UNCorpus:} Michał Ziemski, Marcin Junczys-Dowmunt, and Bruno Pouliquen. 2016. The United Nations Parallel Corpus v1.0. In \textit{Proceedings of the Tenth International Conference on Language Resources and Evaluation (LREC'16)}, pages 3530–3534, Portorož, Slovenia. European Language Resources Association (ELRA).
\item \textbf{translation2019zh:} \url{https://huggingface.co/datasets/yxdu/Translation2019zh}
\item \textbf{WikiMatrix:} Holger Schwenk, Vishrav Chaudhary, Shuo Sun, Hongyu Gong, and Francisco Guzmán. 2021. WikiMatrix: Mining 135M Parallel Sentences in 1620 Language Pairs from Wikipedia. In \textit{Proceedings of the 16th Conference of the European Chapter of the Association for Computational Linguistics: Main Volume}, pages 1351–1361, Online. Association for Computational Linguistics.
\item \textbf{news-commentary:} \url{https://opus.nlpl.eu/News-Commentary/corpus/version/News-Commentary}
\item \textbf{ParaCrawl v9:} \url{https://paracrawl.eu/}
\end{itemize}

Models were checkpointed periodically across training based on both token count and epoch number. For reproducibility, we will release all code, training scripts, and checkpoints alongside this paper.

\end{document}